# LADAR-BASED MOVER DETECTION FROM MOVING VEHICLES

Daniel Morris*, Brian Colonna, Paul Haley
General Dynamics Robotic Systems,
1501 Ardmore Blvd, Pittsburgh, PA 15221

**ABSTRACT**

Detecting moving vehicles and people is crucial for safe operation of UGVs but is challenging in cluttered, real world environments. We propose a registration technique that enables objects to be robustly matched and tracked, and hence movers to be detected even in high clutter. Range data are acquired using a 2D scanning Ladar from a moving platform. These are automatically clustered into objects and modeled using a surface density function. A Bhattacharya similarity is optimized to register subsequent views of each object enabling good discrimination and tracking, and hence mover detection.

## 1. INTRODUCTION

Ladars have been used extensively for real-time mapping and navigation. Stationary obstacles are readily detected and incorporated into a local map, enabling unmanned ground vehicles to plan paths and traverse cluttered environments, for examples see: (Langer et al. 1994; Lacaze et al. 2002; Thrun 2002; Wellington and Stentz 2004). The focus is on detecting and avoiding stationary obstacles. However the most important objects to avoid are other vehicles and people which often move. But moving objects are much more difficult to detect.

The goal of this work is to automatically detect movers, both vehicles and people, using a 2D scanning Ladar on a moving vehicle. The two key components are to find objects in the scene and to analyze their motion. We achieve the first with a simple region-growing clustering of the hits. To perform registration, we develop a technique that models object surfaces with a probability density model. Models are registered and scored by optimizing a similarity measure based on these densities. A discrete implementation enables fast convolution-based registration. The similarity measure is also useful for resolving matching ambiguities and detecting occlusions.

Related work in the area of mover detection includes that of (Biswas et al. 2002; Mertz et al. 2005). Biswas et al. develop a dynamic occupancy grid to model movers, but this is suitable primarily for slow movers in flat, indoor environments. Mertz et al. can detect fast movers outdoors. They group hits from laser line scanners into potential obstacles and track them. The major limitation is that 3D objects are represented by a single slice making both clustering and shape discrimination more difficult. Our work seeks to leverage full 3D surfaces for better discrimination and tracking. In this regard our work is related to 3D model matching techniques such as Iterative Closest Point (ICP) (Besl and McKay 1992) and its variants (Bernardini and Rushmeier 2000; Chen and Medioni 1992; Huber and Hebert 2003; Rusinkewicz and Levoy 2001). ICP and variants suffer from being trapped by local minima unless started close to the correct solution. Also these methods are typically used to register overlapping regions of high-resolution depth maps using six degrees of freedom. Our application is quite different since when movers are at large distances (near the limits of the Ladar), they are sampled very coarsely compared to their curvature and so may appear quite different in successive scans. Also we wish to use constraints such as vehicles move horizontally on the ground. Other registration methods that build a mesh (Bajaj et al. 1995) or assume an interpolated surface between points (Culess and Levoy 1996) provide poor approximations to coarsely sampled points as they may cut off corners or fill gaps between branches or fill an open window. Also, a tree trunk may receive only a single column of hits making a mesh-like surface infeasible. Another weakness of current registration techniques is that they only provide a relative measure for the goodness of a match, and so it is difficult to assess whether the match is valid or not.

Our approach is to build a very general registration technique that does not involve a mesh-like surface assumption and works with very coarsely sampled data. Our technique avoids being trapped by local minima, a problem with ICP, and it provides an absolute measure of goodness to a match enabling discrimination between ambiguities and detections of occlusion or object loss.

## 2. SENSOR DATA

Our sensor is a GDRS Generation IV 2D scanning Ladar with a wide field of view. The scan rate was set at roughly 10Hz. A 2D grid-based depth map is produced which can be projected into 3D as shown in Figure 1(*a*) and (*b*). An onboard INS provides UGV ego-motion and pan-tilt encoders give the sensor-head motion.

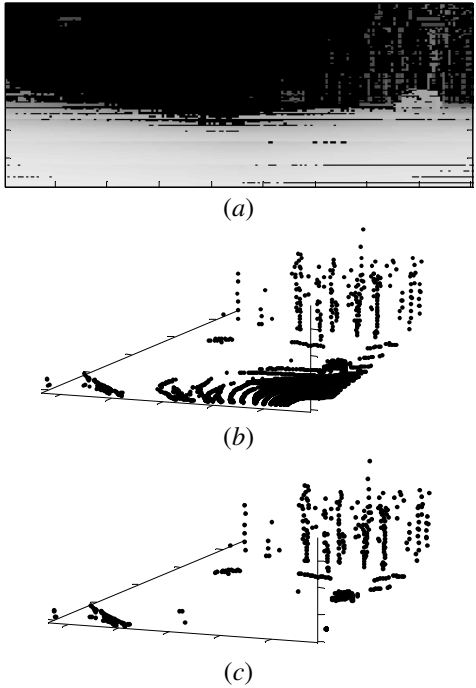

**Figure 1.** (a) Ladar depth map of open terrain containing two vehicles, and trees on the right, (b) resulting 3D projection of points, and (c) 3D points with ground-surface removed.

## 3. APPEARANCE-CHANGE PROBLEM

Detecting movers from a stationary scanning Ladar is straightforward. Beams that are intersected by the mover will change in depth, while the rest remain constant. However, once the Ladar is moving, it is no longer so simple to distinguish movers from stationary objects. This is because Ladar beams sample the angular space with narrow beams at a relatively coarse resolution. A small Ladar motion can produce quite a different sampling of the world space. Figure 2 illustrates a significant appearance change and offset due to small Ladar motion. These sampling effects mean that both movers and stationary objects change appearance between frames.

The mover detection challenge is to distinguish changes due to viewpoint, sampling and occlusions from changes due to target motion. In video, objects are often tracked by matching features, such as color or texture, between images. However, individual Ladar beams are less distinctive as they do not return reflectance properties, and so it is difficult to find features on coarsely sampled objects. Thus our approach is to use whole objects as features and to track these. To achieve this, the following two tasks must be performed: (1) Separate objects must be detected, and (2) These objects must be tracked over time. These are both data association problems.

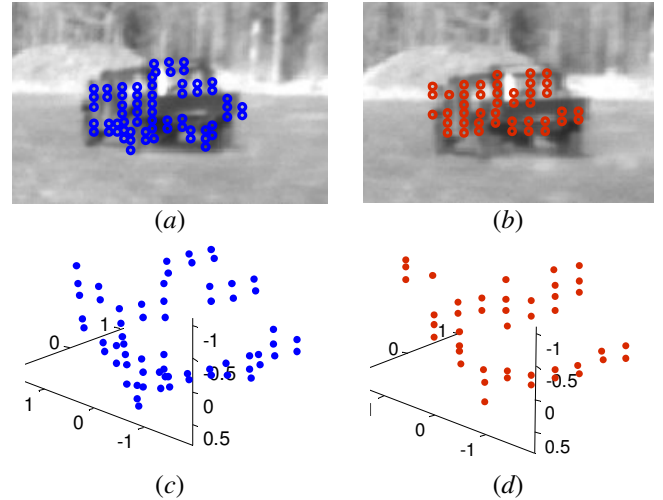

**Figure 2.** Ladar hits on a moving vehicle from subsequent frames in (*a*) and (*b*). Images are for illustration only. Plots (*c*) and (*d*) show corresponding 3D points, illustrating some of the significant effects of sampling on objects.

## 4. OBJECT DETECTION

The first component of our algorithm is to divide the world into separate objects, some of which may be moving. Typically, objects are connected through the ground surface, and so before clustering objects, the ground hits are removed, as illustrated in Figure 1(*c*). A local, roughly horizontal, planar model is used to fit ground points, leaving hits on objects such as trees and vehicles.

The main challenges in clustering hits into objects include the following: (1) the number of objects is unknown, (2) the object sizes can vary greatly, (3) the sampling of objects in the world falls off with the inverse square of distance, and (4) clustering must be done in real time. There are two data-spaces in which clustering can be done. The first is regular *x-y-z* Euclidean space, and the second is angle-depth space, $\theta$–$\phi$–$r$. The advantage of Euclidean space is that it is simple to accumulate stationary data over time, although since our goal is moving-object detection, this does not help much. Angular space has the advantages of having a natural adjacency between hits, as well as a uniform sampling.

Our first approach to clustering was to use mean-shift with a 3D window on the hits, in either Euclidean or angle-depth space. This worked reasonably well, although it was slow in cluttered scenes such as forests. It could be sped up to real time by doing only incremental clustering between frames. However, the main drawback is the dependency of clusters on the mean-shift window size. Objects much larger than the window tended to break into multiple clusters. For example, when person-sized windows were used, a wall would be broken up into roughly person-sized clusters. This posed problems for the next step of temporal association of clusters, sometimes resulting in ambiguous matching and false positives. An additional step of clustering the close-by clusters can remediate this, but this adds complexity.

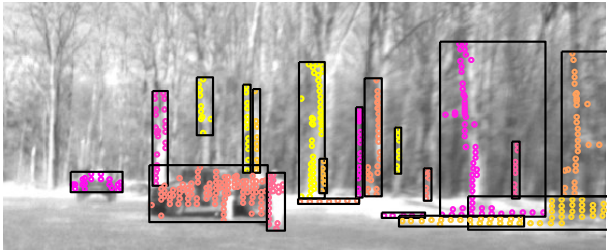

**Figure 3**. Example of clustering objects at long range. Ladar hits are plotted on top of the image for ease of viewing only. Trees are found in the background, two vehicles on the left, a person in the center-front and some brush at the right.

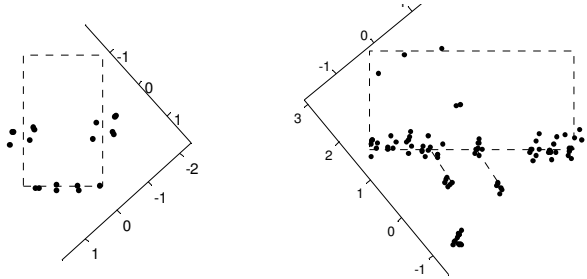

**Figure 4**. Top-down views of the two vehicles clustered in Figure 3. The dashed rectangles show the true location of the vehicle. Notice the vehicle on the right has two doors open and a stationary close-by person is clustered with it.

The technique that worked the best was to do contiguous region building. This leverages the adjacency information in the angle-depth space, has very low computational requirements, and works well for both large and small objects. A depth threshold, $d_T$, was defined such that adjacent points separated by less than $d_T$ were clustered together using region growing. Note that since the tangential distance between sampled points grows with depth, this threshold was made proportional to depth.

## 5. OBJECT REGISTRATION

The most difficult task, and hence the core of our work, is registering objects between frames. Consider for now a single object being viewed in multiple subsequent frames. Our goal is to determine if it is a mover, and if so what its velocity is. The simplest approach would be to analyze the motion of the centroid of the points. This has two potential pitfalls. First, if different portions of the surface are sampled in different frames due to partial occlusions or self-occlusions, the centroid could move significantly leading to false motion estimates. Second, without an object similarity measure it is possible for the incorrect object to be matched and again false motions obtained. Hence we desire a more precise and more discriminating registration technique.

In each frame, $f$, a set of points in world coordinates lying on the visible surface of the object surface is obtained, $^fX = \{x_1,\ldots,x_n\}$. The number of points in each frame can vary depending on viewing angle. Our registration approach relies on explicitly modeling the object surface as a probability density function, $\rho_S(^fX)$, given the set of 3D sampled points. If the sampling density is high compared to the surface curvature, then a mesh might be a good representation, however in our case the opposite is true; the sampling for objects at long range can be very coarse and so the surface between samples can vary significantly or may have gaps. Hence we use a mixture of 3D Gaussians centered at each sampled point for our density function:

$$\rho_S(^fX) = \sum_i N(x_i, \sigma_i^2)/n. \qquad (1)$$

The covariances $\sigma_i^2$ are proportional to the sampling density, and hence to the distance from the Ladar. This models a wide variety of surfaces including coarsely sampled natural objects such as trees.

Now as both the object and the Ladar move, different points on its surface will be sampled. Denote the points in the next frame as $^gX$, and its new surface density as: $\rho_S(^gX)$. Also denote the object motion between frames $f$ and $g$ as $^g_fT$ and its inverse as $^f_gT$ such that $^f_gT\,^gX$ maps points in frame $g$ to their corresponding positions in frame $f$. If the same points on the object were sampled in both frames, then the transformed density: $\rho_S\left({}_g^f T {}^g X\right)$ would exactly equal $\rho_S\left({}^f X\right)$. The object motion could be estimated by finding the transformation, ${}_g^f T$, that achieves this. In general different object points will be sampled and so the densities will not be exactly equal. Hence we need a similarity measure between densities that can be optimized as a function of ${}_g^f T$. The Bhattacharya similarity provides such a measure, although other measures can be used too:

$$B\left({}_g^f T\right) = \int_x \sqrt{\rho_S\left({}_g^f T {}^g X\right) \rho_S\left({}^f X\right)}. \qquad (2)$$

It has a range of 0 to 1, and reaches 1 when the two densities are equal. Optimizing Equation (2) as a function of ${}_g^f T$ gives the best shape match, see Figure 6. Also the value $B\left({}_g^f T\right)$ is a useful absolute measure of how well the two surfaces match. This is an important advantage over other 3D registration techniques such as ICP which only give relative goodness of matches. A low value of $B\left({}_g^f T\right)$ can indicate an occlusion or an incorrect match, and so it is useful when there are multiple matches to choose from.

Since some frames may contain poor views of an object, and since the viewpoint changes over time, we accumulate object surface densities over time. After each subsequent frame is registered to the current model, it is appropriately transformed and added to the model. In this way a higher density surface is gradually created. Since errors may accumulate in registration, and because the most recent past is the most useful for future registration, we decay the weighting of old measurements with an appropriate half-life.

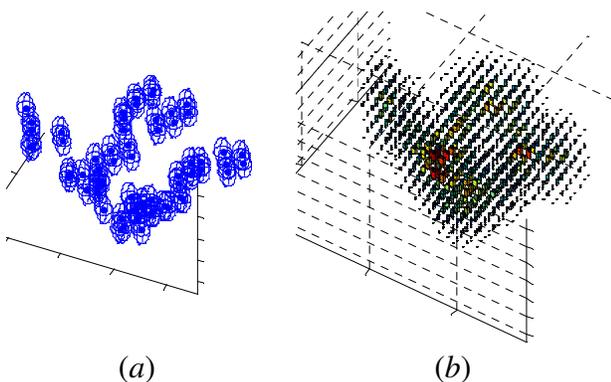

(a) (b)

**Figure 5**. (*a*) A surface density model represented as a mixture of Gaussians $\rho_S\left({}^f X\right)$ on the object in Figure 2(*c*). On the right a discretization of this model, $\rho_{DS}\left({}^f X\right)$, into a 29x29x7 grid.

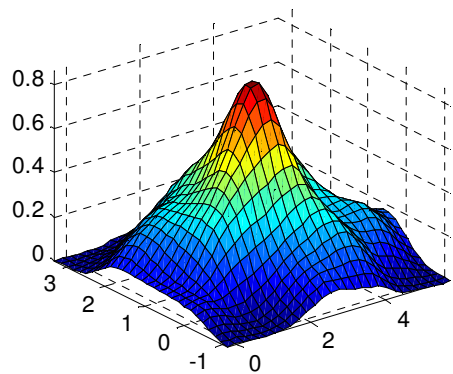

**Figure 6**. The Bhattacharya similarity, $B_D\left({}_g^f T\right)$, of the discretized density surfaces as shown in Figure 5 for the vehicle views in Figure 2. A parabolic fit at the peak gives an optimum $B_D\left({}_g^f T\right)$ of 0.88. Here only translation is modeled; including rotation would add an extra dimension to the plot. In this case there is a single maximum, but there may be multiple maxima when viewpoints vary more.

For real-time operation we developed a discrete implementation. The density functions are binned into a 3D grid, $\rho_{DS}\left({}^f X\right)$, as illustrated in Figure 5. Then the discrete Bhattacharya similarity $B_D\left({}_g^f T\right)$ for a range of motions is found as the convolution of the square root of the densities:

$$B_D\left({}_g^f T\right) = \sqrt{\rho_{DS}\left({}_g^f T {}^g X\right)} * \sqrt{\rho_{DS}\left({}^f X\right)}. \qquad (3)$$

When motion is assumed to be translation in the horizontal plane, these 3D convolutions are simply the sum of the 2D horizontal-slice convolutions, see Figure 6. The best motion estimate, ${}_g^f T$, can be found at the maximum of this surface. Sub-grid-size precision is obtained by local parabolic fitting around the maximum.

A problem facing multi-object tracking algorithms is the high variability in computational load, especially when there are many objects and clutter in the scene. A significant advantage of discretizing the density function is that it gives great flexibility in adjusting the computational load. Density functions can be smoothed and sampled at reduced resolution to reduce computation when needed. In particular, reducing resolution in the vertical dimension gives speed gains with little loss of tracking accuracy. When computation was scarce, we found that reducing vertical resolution to three slices gave a good speedup while maintaining good robustness to partial occlusions. Sampling only a single slice could reduce computation further, but we found it more susceptible to incorrect matches caused by partial occlusions.

For many objects and motions only a horizontal translation model is sufficient. This includes objects that are roughly rotationally invariant such as people that at long range appear like vertical cylinders, or stationary objects such as trees, and even vehicles when their angular motion is gradual. The approximation errors are compensated for by the decaying model coupled with the tracker. However, when vehicles turn sharply, rapid density model change can cause a loss of registration. To account for these cases it is useful to model rotations around the vertical axis. A straightforward extension to achieve this is to discretize the latest surface density at several rotations. Each of these is convolved with the accumulated model, and a layered convolution surface is obtained. The maximum location of this can be found by fitting a parabola in 3 dimensions around the peak. Since rotation between frames is typically not great, we found it sufficient to sample between 3 and 5 rotations.

## 6. OBJECT TRACKING

Most of the work in moving object detection is achieved by clustering and registration. However, there are a number of sources of clutter as well as objects appearing and disappearing due to occlusions. These effects can lead to spurious motion estimates and hence false positives. To minimize these we enforced motion consistency using a Kalman filter tracker. All objects, including stationary objects, are tracked as long as their motion does not exceed an acceleration limit. Those objects whose speed exceeds a minimum threshold are declared to be movers. The Kalman filter gives a predicted location and uncertainty for each object in a new frame, and this is used to bound the search region during registration. Since the frame rate is high there is typically no ambiguity in cluster association. But sometimes in high clutter or after an occlusion there are multiple potential matches, and we select the best one using the density similarity function in Equation (3).

## 7. RESULTS

Figure 7 contains a short sequence illustrating moving vehicle detection from a moving UGV. The algorithm runs in real-time on a Pentium processor.

Our algorithm was tested on a number of ground-truthed runs in a variety of environments including wooded, open and urban environments as shown in Figure 8. The hit percentage per target shows the percent of frames in which each mover was correctly identified as a mover. This is lower in high clutter environments due to occlusions. A consistently low false alarm rate was maintained.

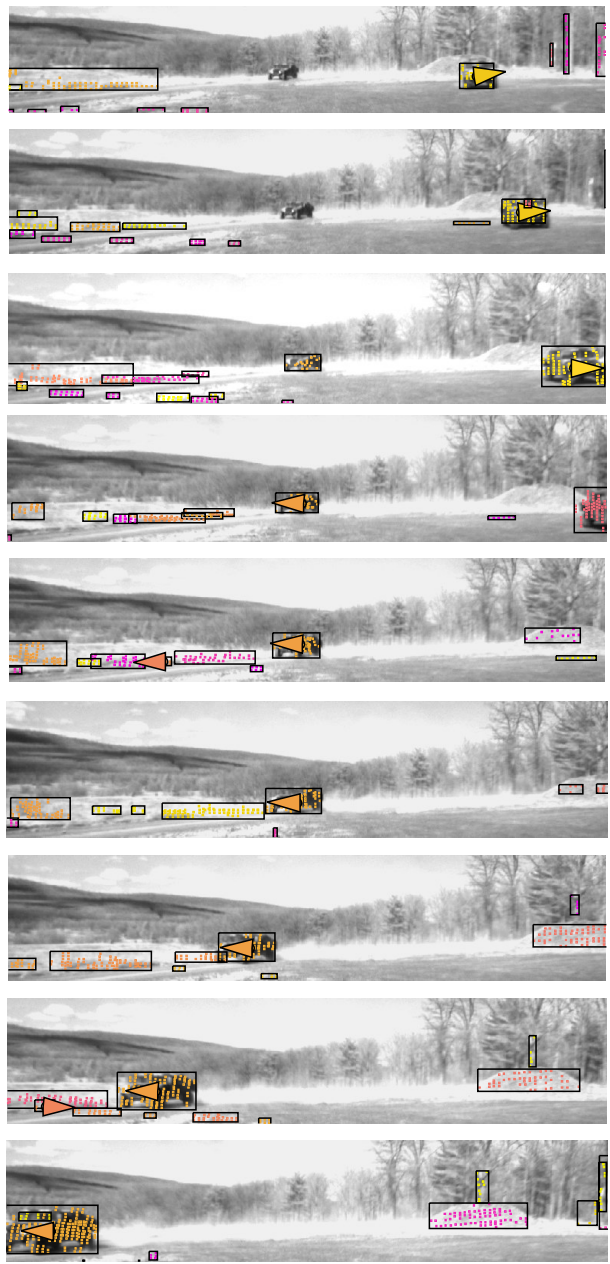

**Figure 7.** A portion of the Ladar field of view containing every 4$^{th}$ frame of a sequence with two moving vehicles. Clusters are shown by their bounding boxes and those identified as movers are marked with arrows. In the top two images the center vehicle is beyond the range hits, while the vehicle to the right is being tracked. In the 3$^{rd}$ image the center vehicle is detected and by the 4$^{th}$ it is identified as a mover. The ground clutter on the left occasionally generates false positives.

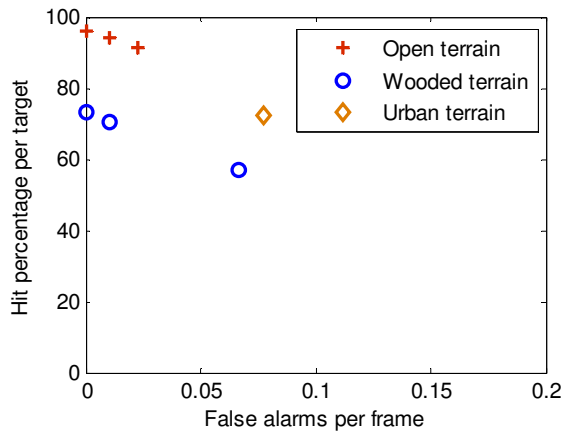

**Figure 8**. Detection performance using ground-truthed data runs. The hit percentage per target is lower for wooded terrain because of numerous occlusions. This is because misses are accumulated after each occlusion while the algorithm determines it is a mover. The false alarm rates are consistently low and slightly higher with more clutter.

## CONCLUSION

We have developed a surface probability density model for 3D object registration. It does not require a mesh and is appropriate for objects that are coarsely sampled. The Bhattacharya measure comparing two density functions gives an absolute similarity estimate between 0 and 1, enabling the goodness of a match to be assessed. Our discrete implementation using convolution filtering enables real-time registration without being trapped by local minima. Horizontal translation and rotation about the vertical axis are modeled. When integrated with a tracker, we obtained a robust mover detector that can handle high clutter and significant self-motion, and still obtain high detection rates with few false alarms.

## ACKNOWLEDGEMENTS:

This work was sponsored by the Army Research Lab under contract: DAAD 19-01-2-0012